\documentclass[11pt,a4paper]{article}
\usepackage[hyperref]{acl2020}
\usepackage{times}
\usepackage{latexsym}

\usepackage{microtype}

\usepackage{url}
\usepackage{amsfonts} 
\usepackage{booktabs} 
\usepackage{multirow} 
\usepackage{subfigure} 
\usepackage{xspace}
\usepackage{graphicx}
\usepackage{color}
\usepackage{colortbl}
\usepackage[english]{babel}
\usepackage{balance}
\usepackage{amsmath,bm}
\usepackage{makecell}
\usepackage{algorithm}
\usepackage{algpseudocode}

\definecolor{Gray}{gray}{0.9}

\aclfinalcopy 
\def\aclpaperid{3066} 

\title{Neighborhood Matching Network for Entity Alignment}

\author{
	Yuting Wu$^{1}$, 
	Xiao Liu$^{1}$, 
	Yansong Feng$^{1,2}$\thanks{\;\;Corresponding author.},  
	Zheng Wang$^{3}$ \and
	Dongyan Zhao$^{1,2}$ \\
	$^1$Wangxuan Institute of Computer Technology, Peking University, China\\
	$^2$The MOE Key Laboratory of Computational Linguistics, Peking University, China\\
	$^3$School of Computing, University of Leeds, U.K. \\
	{\tt \{wyting,lxlisa,fengyansong,zhaodongyan\}@pku.edu.cn} \\
	{\tt z.wang5@leeds.ac.uk}\\
}

\date{}

\begin{document}
\maketitle

\begin{abstract}
Structural heterogeneity between knowledge graphs is an outstanding challenge for entity alignment. This paper presents Neighborhood Matching Network (NMN), a novel entity alignment framework for tackling the structural heterogeneity challenge. NMN estimates the similarities between entities to capture both the topological structure and the neighborhood difference. It provides two innovative components for better learning representations for entity alignment. It first uses a novel graph sampling method to distill a discriminative neighborhood for each entity. It then adopts a cross-graph neighborhood matching module to jointly encode the neighborhood difference for a given entity pair. Such strategies allow NMN to effectively construct matching-oriented entity representations while ignoring noisy neighbors that have a negative impact on the alignment task. Extensive experiments performed on three entity alignment datasets show that NMN can well estimate the neighborhood similarity in more tough cases and significantly outperforms 12 previous state-of-the-art methods.

\end{abstract}
	
\section{Introduction}
\label{section:intro}
By aligning entities from different knowledge graphs (KGs) to the same real-world identity, \emph{entity alignment} is
a powerful technique for knowledge integration. Unfortunately, entity alignment is non-trivial because real-life KGs are often
incomplete and different KGs typically have heterogeneous schemas. Consequently, equivalent entities from two KGs could
have distinct surface forms or dissimilar neighborhood structures.

\begin{figure}
	\centering
	\includegraphics[width=0.85\linewidth]{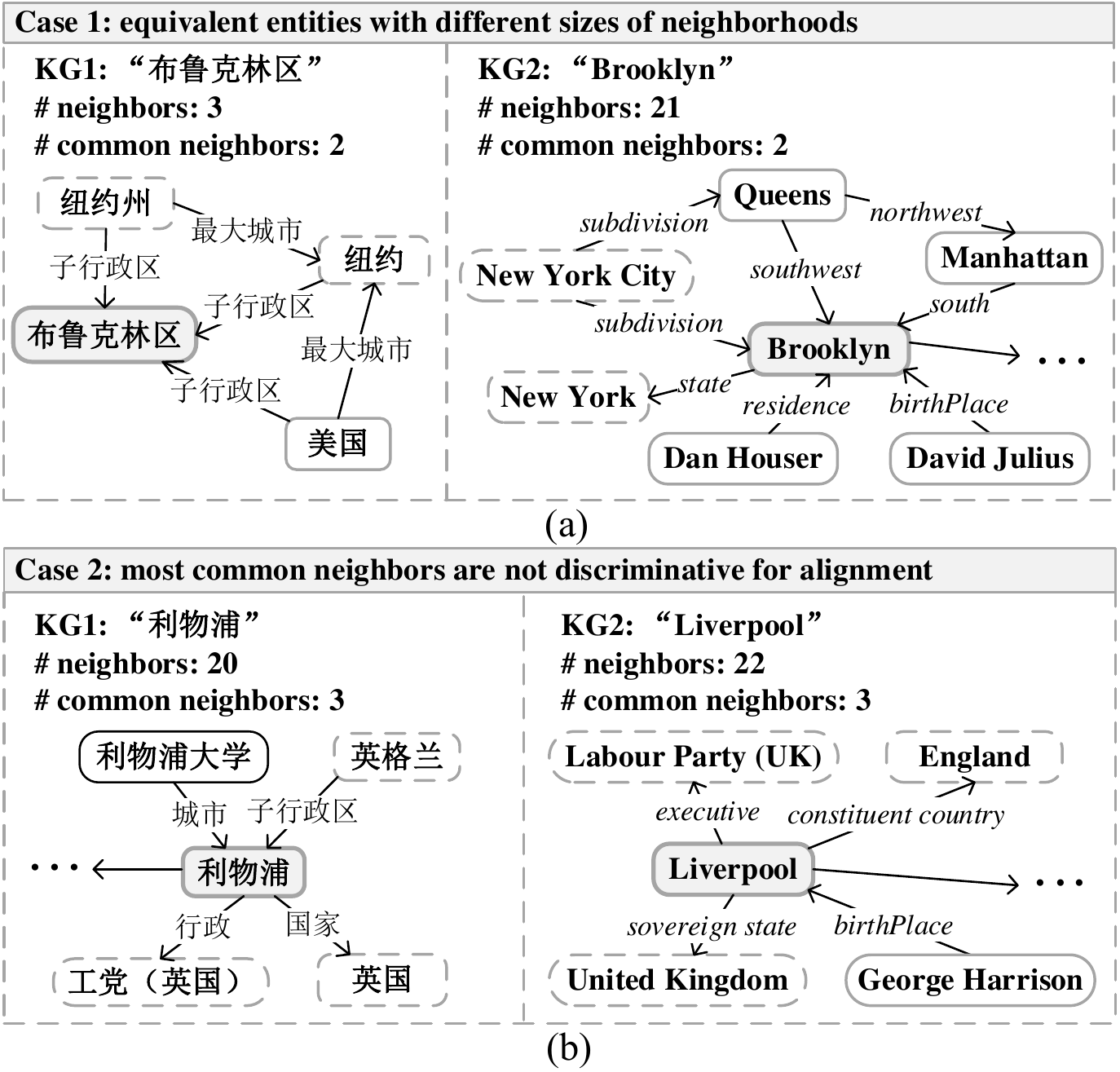}
	\vspace{-2mm}
    \caption{Illustrative examples: two tough cases for entity alignment. Dashed rectangles denote the common neighbors between different KGs.}
	\label{case}
	\vspace{-3mm}
\end{figure}

In recent years, embedding-based methods have become the dominated approach for entity
alignment~\citep{zhu2017iterative,pei2019semi,cao2019multi,gmnn_acl19,li2019semi,sun2019knowledge}. Such approaches have the advantage of
not relying on manually constructed features or rules~\citep{Mahdisoltani2014YAGO3AK}. Using a set of seed alignments, an embedding-based
method models the KG structures to  automatically learn how to map the equivalent entities among different KGs into a unified vector space
where entity alignment can be performed by measuring the distance between the embeddings of two entities.

The vast majority of prior works in this direction build upon an important assumption - entities and their counterparts
from other KGs have similar neighborhood structures, and therefore, similar embeddings will be generated for equivalent
entities. Unfortunately, the assumption does not always hold for real-life scenarios due to the incompleteness and
heterogeneities of KGs. As an example, consider Figure~\ref{case}~(a), which shows two equivalent entities from the
Chinese and English versions of Wikipedia. Here, both central entities refer to the same real-world identity,
\emph{Brooklyn}, a borough of New York City. However, the two entities have different sizes of neighborhoods and
distinct topological structures. The problem of dissimilar neighborhoods between equivalent entities is ubiquitous.
\citet{sun2019knowledge} reports that the majority of equivalent entity pairs have different neighbors in the benchmark
datasets DBP15K, and the proportions of such entity pairs are over 86\% (up to 90\%) in different language versions of
DBP15K. Particularly, we find that the alignment accuracy of existing embedding-based methods decreases significantly
as the gap of equivalent entities' neighborhood sizes increases. For instance, RDGCN
\citep{Wu:2019:REA:3367722.3367779}, a state-of-the-art, delivers an accuracy of 59\% on the Hits@1 score on entity
pairs whose number of neighbors differs by no more than 10 on DBP15K$_{ZH-EN}$. However, its performance drops
to 42\% when the difference for the number of neighbors increases to 20 and to 35\% when the difference increases to be
above 30. The disparity of the neighborhood size and topological structures pose a significant challenge for entity
alignment methods.

Even if we were able to set aside the difference in the neighborhood size, we still have another issue. Since most of
the common neighbors would be popular entities, they will be neighbors of many other entities. As a result, it is still
challenging to align such entities. To elaborate on this point, let us now consider Figure~\ref{case}~(b). Here, the
two central entities (both indicate the city \emph{Liverpool}) have similar sizes of neighborhoods and three common
neighbors. However, the three common neighbors (indicate \emph{United Kingdom}, \emph{England} and \emph{Labour Party
(UK)}, respectively) are not discriminative enough. This is because there are many \emph{city} entities for England which also have the three entities in their neighborhoods -- e.g., the entity \emph{Birmingham}. For such entity pairs, in
addition to common neighbors, other informative neighbors -- like those closely contextually related to the central
entities -- must be considered. Because existing embedding-based methods are unable to choose the right neighbors, we need a better approach.

We present Neighborhood Matching Network (NMN), a novel sampling-based entity alignment framework. NMN aims to capture the most informative neighbors and accurately estimate the similarities of neighborhoods between
entities in different KGs. NMN achieves these by leveraging the recent development in Graph Neural Networks (GNNs). It
first utilizes the Graph Convolutional Networks (GCNs) \citep{Kipf2016Semi} to model the topological connection
information, and then selectively samples each entity's neighborhood, aiming at retaining the most informative
neighbors towards entity alignment. One of the key challenges here is how to accurately estimate the similarity of any
two entities' sampled neighborhood. NMN addresses this challenge by designing a discriminative
neighbor matching module to jointly compute the neighbor differences between the sampled subgraph pairs through a cross-graph attention mechanism.
Note that we mainly focus on the neighbor relevance in the neighborhood sampling and matching modules, while the neighbor connections are modeled by GCNs. We show that, by integrating the neighbor connection information and the neighbor relevance information, NMN can effectively
align entities from real-world KGs with neighborhood heterogeneity.

We evaluate NMN by applying it to benchmark datasets DBP15K \citep{sun2017cross} and DWY100K \citep{ijcai2018-611}, and a sparse variant of DBP15K. 
Experimental results show that NMN achieves the best and more robust performance over state-of-the-arts.
This paper makes the following technical contributions. It is the first to:
\begin{itemize}
\item employ a new graph sampling strategy for identifying the most informative neighbors towards entity alignment (Sec.~\ref{neighbor_sample}). 

\item exploit a cross-graph attention-based matching mechanism to jointly compare discriminative subgraphs of two
entities for robust entity alignment (Sec.~\ref{neighbor_match}).

\end{itemize}

\section{Related Work}
\paragraph{Embedding-based entity alignment.} \label{related-EA} 
In recent years, embedding-based methods have emerged as viable means for entity alignment. Early works in the area utilize
TransE \citep{bordes2013translating} to embed KG structures, including MTransE
\citep{chen2016multilingual}, JAPE \citep{sun2017cross}, IPTransE \citep{zhu2017iterative}, BootEA \citep{ijcai2018-611}, NAEA \citep{zhu2019neighborhood} and OTEA \citep{pei2019improving}. Some more recent studies use GNNs to model the
structures of KGs, including GCN-Align \citep{D18-1032}, GMNN \citep{gmnn_acl19}, RDGCN
\citep{Wu:2019:REA:3367722.3367779}, AVR-GCN \citep{ye2019vectorized}, and HGCN-JE \citep{wu2019jointly}. Besides the structural information, some recent methods like KDCoE \citep{Chen2018Co}, AttrE \citep{trisedya2019entity}, MultiKE \cite{MultiKE} and HMAN \citep{yang2019aligning} also utilize additional information like Wikipedia entity descriptions and attributes to improve entity representations.

However, all the aforementioned methods ignore the neighborhood heterogeneity of KGs. MuGNN \citep{cao2019multi} and AliNet
\citep{sun2019knowledge} are two most recent efforts for addressing this issue. While promising, both models still have drawbacks.
MuGNN requires both pre-aligned entities and relations as training data, which can have expensive overhead for training data labeling.
AliNet considers all one-hop neighbors of an entity to be equally important when aggregating information. However, not all one-hop neighbors contribute positively to characterizing the target entity. Thus, considering all of them without careful selection can introduce noise and degrade the performance. NMN avoids these pitfalls. With only a small set of pre-aligned entities as training data, NMN chooses the most informative neighbors for entity alignment.

\paragraph{Graph neural networks.}

GNNs have recently been employed for various NLP tasks like semantic role labeling \citep{Marcheggiani2017Encoding} and machine translation
\citep{Bastings2017Graph}. GNNs learn node representations by recursively aggregating the representations of neighboring nodes. There are a range of GNN variants, including the Graph Convolutional Network (GCN) \citep{Kipf2016Semi}, the Relational Graph Convolutional Network \citep{schlichtkrull2018modeling}, the Graph Attention Network \citep{velickovic2018graph}. 
Giving the powerful capability for modeling graph structures, we also leverage GNNs to encode the structural information of KGs~(Sec.~\ref{structure-embed}).

\paragraph{Graph matching.}
The similarity of two graphs can be measured by exact matching (graph isomorphism) \citep{yan2004graph} or through
structural information like the graph editing distance \citep{raymond2002rascal}. Most recently, the Graph Matching
Network (GMN) \citep{li2019graph} computes a similarity score between two graphs by jointly reasoning on the graph pair
through cross-graph attention-based matching. Inspired by GMN, we design a cross-graph neighborhood matching module
(Sec. \ref{neighbor_match}) to capture the neighbor differences between two entities' neighborhoods.

\paragraph{Graph sampling.} This technique samples a subset of vertices or edges from the original graph. Some of the popular sampling approaches
include vertex-, edge- and traversal-based sampling \citep{hu2013survey}. In our entity alignment framework, we
propose a vertex sampling method to select informative neighbors and to construct a neighborhood subgraph for each entity.

\section{Our Approach}
\begin{figure*}
    \centering
	\includegraphics[width=0.7\linewidth]{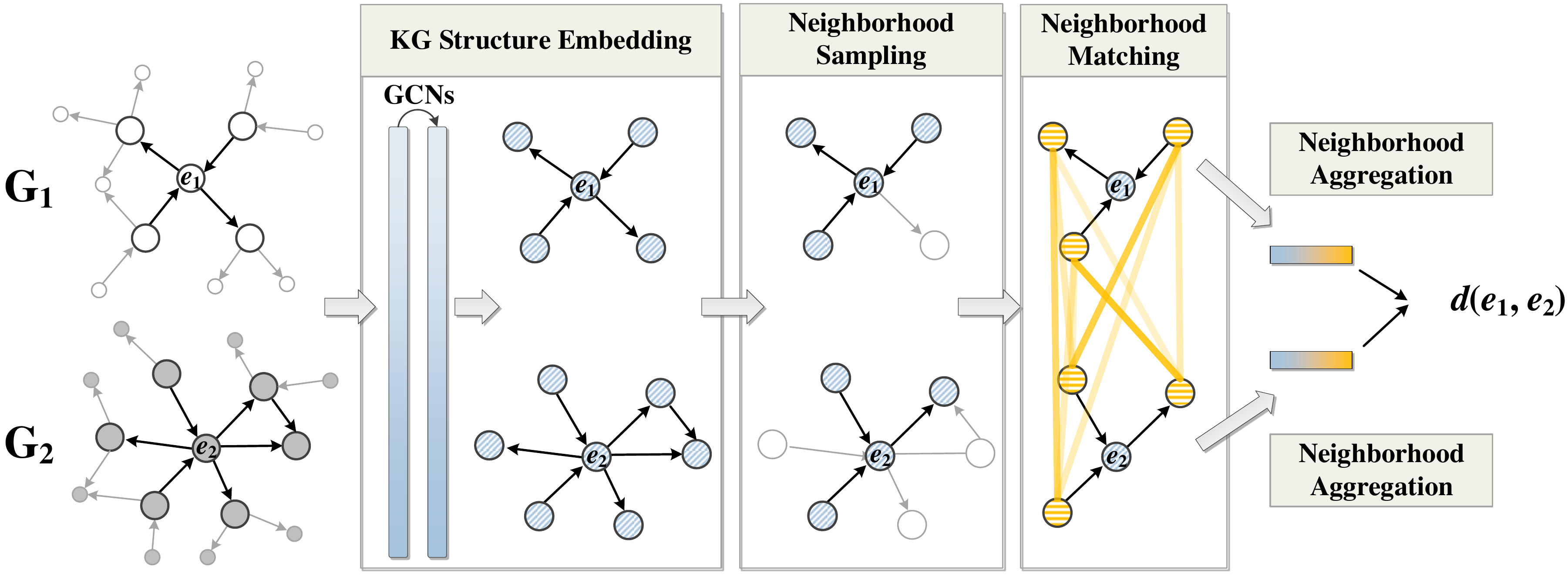}
	\vspace{-2mm}
	\caption{Overall architecture and processing pipeline of Neighborhood Matching Network (NMN).}
	\label{overall}
	\vspace{-4mm}
\end{figure*}
Formally, we represent a KG as $G=(E,R,T)$, where $E,R,T$ denote the sets of entities, relations and triples respectively. Without loss of generality, we consider the task of entity alignment between two KGs, $G_1$ and $G_2$, based on a set of pre-aligned equivalent entities.
The goal is to find pairs of equivalent entities between $G_1$ and $G_2$.
\subsection{Overview of NMN}
\label{sec:overview}

As highlighted in Sec.~\ref{section:intro}, the neighborhood heterogeneity and noisy common neighbors of real-world KGs make it difficult to
capture useful information for entity alignment. To tackle these challenges, NMN first leverages GCNs to model the neighborhood
topology information. Next, it employs neighborhood sampling to select the more informative neighbors. Then, it utilizes a cross-graph matching module to capture neighbor differences.

As depicted in Figure \ref{overall}, NMN takes as input two KGs, $G_1$ and $G_2$, and produces embeddings for each candidate pair of entities, $e_1$
and $e_2$, so that entity alignment can be performed by measuring the distance, $d(e_1, e_2)$, of the learned embeddings. It follows
a four-stage processing pipeline: (1) KG structure embedding, (2) neighborhood sampling, (3) neighborhood matching, and (4) neighborhood
aggregation for generating embeddings.

\subsection{KG Structure Embedding}
\label{structure-embed} To learn the KG structure embeddings, NMN utilizes multi-layered GCNs to aggregate higher degree neighboring structural information for
entities.

NMNs uses pre-trained word embeddings to initialize the GCN. This strategy is shown to be effective in encoding the semantic information of
entity names in prior work \citep{gmnn_acl19,Wu:2019:REA:3367722.3367779}.
 Formally, let $G_1=(E_1,R_1,T_1)$ and
$G_2=(E_2,R_2,T_2)$ be two KGs to be aligned, we put $G_1$ and $G_2$ together as one big input graph to NMN. Each GCN layer
takes a set of node features as input and updates the node representations as:

\begin{equation}
    \bm{h}_i^{(l)}={\rm ReLU}(\sum\limits_{j \in N_i \cup \{i\}} \frac{1}{\epsilon_i}\textbf{W}^{(l)}\bm{h}_j^{(l-1)})
\end{equation}
where $\{\bm{h}^{(l)}_1,\bm{h}^{(l)}_2,...,\bm{h}^{(l)}_{n} |\bm{h}^{(l)}_{i} \in \mathbb{R}^{d^{(l)}}\}$ is the output node (entity)
features of $l$-th GCN layer, $\epsilon_i$ is the normalization constant, $N_i$ is the set of neighbor indices of entity $i$, and
$\textbf{W}^{(l)} \in \mathbb{R}^{d^{(l)}\times d^{(l-1)}}$ is a layer-specific trainable weight matrix.

To control the accumulated noise,  we also introduce highway networks \citep{Srivastava2015Highway} to GCN layers, which can effectively control the noise propagation across GCN layers~\citep{Rahimi2018Semi,wu2019jointly}.

\subsection{Neighborhood Sampling}
\label{neighbor_sample} 
The one-hop neighbors of an entity are key to determine whether the entity should
be aligned with other entities. However, as we have discussed in Sec.~\ref{section:intro}, not all one-hop neighbors
contribute positively for entity alignment. To choose the right neighbors, we apply a down-sampling process to select
the most informative entities towards the central target entity from its one-hop neighbors.

Recall that we use pre-trained word embeddings of entity names to initialize the
input node features of GCNs. As a result, the entity embeddings learned by GCNs contain rich contextual information for both the neighboring
structures and the entity semantics. NMN exploits such information to  sample informative neighbors, i.e., neighbors that are more contextually related to the
central entity are more likely to be sampled. Our key insight is that the more often a neighbor and the central (or target) entity appear
in the same context, the more representative and informative the neighbor is towards the central entity. Since the contexts of two
equivalent entities in real-world corpora are usually similar, the stronger a neighbor is \emph{contextually related} to the target entity,
the more alignment clues the neighbor is likely to offer. Experimental results in Sec. \ref{analyses} confirm this observation.

Formally, given an entity $e_i$, the probability to sample its one-hop neighbor $e_{i\_j}$ is determined by:
\begin{equation}
\label{sample_distribution}
\begin{split}
    p(\bm{h}_{i\_j}|\bm{h}_i)&={\rm softmax}(\bm{h}_i \textbf{W}_s \bm{h}_{i\_j}^T)\\
&=\frac{exp(\bm{h}_i \textbf{W}_s \bm{h}_{i\_j}^T)}{\sum_{k \in N_i}exp(\bm{h}_i \textbf{W}_s \bm{h}_{i\_k}^T)}
\end{split}
\end{equation}
where $N_i$ is the one-hop neighbor index of central entity $e_i$, $\bm{h}_i$ and $\bm{h}_{i\_j}$ are learned
embeddings for entities $e_i$ and $e_{i\_j}$ respectively, and $\textbf{W}_s$ is a shared weight matrix.

By selectively sampling one-hop neighbors, NMN essentially constructs a discriminative subgraph of neighborhood for each entity, which can enable more accurate alignment through neighborhood matching.

\subsection{Neighborhood Matching}
\label{neighbor_match} 
The neighborhood subgraph, produced by the sampling process, determines which neighbors of the target entity should be considered in the
later stages. In other words, later stages of the NMN processing pipeline will only operate on neighbors within the subgraph. In the
neighborhood matching stage, we wish to find out,
for each candidate  entity in the counterpart KG, which neighbors of that entity are
closely related to a neighboring node within the subgraph of the target entity.
Such information is essential for deciding whether two
entities (from two KGs) should be aligned.

As discussed in Sec.~\ref{neighbor_sample}, equivalent entities tend to have similar contexts in real-world corpora; therefore, their
neighborhoods sampled by NMN should be more likely to be similar. NMN exploits this observation to estimate the similarities of the sampled neighborhoods.

\paragraph{Candidate selection.}
Intuitively, for an entity $e_i$ in $E_1$, we need to compare its sampled neighborhood subgraph with the subgraph of each candidate entity
in $E_2$ to select an optimal alignment entity. Exhaustively trying all possible entities of $E_2$  would be prohibitively expensive for large real-world KGs. To reduce the matching overhead, NMN takes a low-cost approximate approach. To that end, NMN first samples an
alignment candidate set $\mathcal{C}_i=\{c_{i_1},c_{i_2},...,c_{i_t}| c_{i_k} \in E_2\}$ for $e_i$ in $E_1$, and then calculates the
subgraph similarities between $e_i$ and these candidates. This is based on an observation that the entities in $E_2$ which are closer to
$e_i$ in the embedding space are more likely to be aligned with $e_i$. Thus, for an entity $e_j$ in $E_2$, the probability that it is
sampled as a candidate for $e_i$ can be calculated as:
\begin{equation}
\label{sample_candidate}
    p(\bm{h}_{j}|\bm{h}_i)=\frac{exp(\|\bm{h}_i-\bm{h}_j\|_{L_1})}{\sum_{k \in E_2}exp(\|\bm{h}_i-\bm{h}_k\|_{L_1})}
\end{equation}

\paragraph{Cross-graph neighborhood matching.}
Inspired by recent works in graph matching~\citep{li2019graph}, our neighbor matching module takes a pair of subgraphs as input, and computes a cross-graph
matching vector for each neighbor, which measures how well this neighbor can be matched to any neighbor node in the counterpart. Formally, let
$(e_i,c_{i_k})$ be an entity pair to be measured, where $e_i \in E_1 $ and $c_{i_k} \in E_2$ is one of the candidates of $e_i$, $p$
and $q$ are two neighbors of $e_i$ and $c_{i_k}$, respectively. The cross-graph matching vector for neighbor $p$ can be computed as:
\begin{equation}
a_{pq}=\frac{{\rm exp}(\bm{h}_p \cdot \bm{h}_q)}{\sum_{q' \in N_{i_k}^s}{\rm exp}(\bm{h}_p \cdot \bm{h}_{q'})}
\end{equation}
\vspace{-2mm}
\begin{equation}
\bm{m}_p=\sum_{q\in N_{i_k}^s} a_{pq}(\bm{h}_p-\bm{h}_q)
\end{equation}
where $a_{pq}$ are the attention weights, $\bm{m}_p$ is the matching vector for $p$, and it measures the difference between $\bm{h}_p$ and
its closest neighbor in the other subgraph, $N_{i_k}^s$ is the sampled neighbor set of $c_{i_k}$, $\bm{h}_p$ and $\bm{h}_q$ are the
GCN-output embeddings for $p$ and $q$ respectively.

Then, we concatenate neighbor $p$'s GCN-output embeddings with weighted matching vector $\bm{m}_p$:
\begin{equation}
\hat{\bm{h}}_p=[\bm{h}_p \| \beta*\bm{m}_p]
\end{equation}

For each target neighbor in a neighborhood subgraph, the attention mechanism in the matching module can accurately detect which of the
neighbors in the subgraph of another KG is most likely to match the target neighbor. Intuitively, the matching vector $\bm{m}_p$ captures
the difference between the two closest neighbors. When the representations of the two neighbors are similar, the matching vector tends to
be a zero vector so that their representations stay similar. When the neighbor representations differ, the matching vector will be
amplified through propagation.
We find this matching strategy works well for our problem settings.

\subsection{Neighborhood Aggregation}
\label{neighborhood-agg}
In the neighborhood aggregation stage, we combine the neighborhood connection information (learned at the KG structure embedding stage) as well as the output of the matching stage (Sec.~\ref{neighbor_match}) to generate the final embeddings
used for alignment.

 Specifically, for entity $e_i$, we first aggregate its sampled neighbor representations $\{\hat{\bm{h}}_p\}$. Inspired by the aggregation method in~\citep{li2015gated}, we compute a neighborhood representation for $e_i$ as:
 \begin{equation}
 \label{neighborhood-level}
     \bm{g}_i=(\sum_{p\in N_i^s}\sigma(\hat{\bm{h}}_p \textbf{W}_{gate})\cdot \hat{\bm{h}}_p)\textbf{W}_N
     \vspace{-2mm}
 \end{equation}

 Then, we concatenate the central entity $e_i$'s GCN-output representation $\bm{h}_i$ with its neighborhood representation to construct the matching oriented representation for $e_i$:
  \begin{equation}
     \bm{h}^{match}_i=[\bm{g}_i \| \bm{h}_i]
 \end{equation}

\subsection{Entity Alignment and Training}\label{sec:alignment}
\paragraph{Pre-training.} As discussed in Sec.~\ref{neighbor_sample}, our neighborhood sampling is based on the GCN-output entity embeddings. Therefore, we first pre-train the GCN-based KG embedding model to produce quality entity representations.
Specifically, we measure the distance between two entities to determine whether they should be
aligned:
\begin{equation}
\label{distance}
\tilde{d}(e_1,e_2)=\|\bm{h}_{e_1}-\bm{h}_{e_2}\|_{L_1}
\end{equation}
The objective of the pre-trained model is:
\begin{equation}
\small
\tilde{L}=\sum\limits_{(i,j)\in \mathbb{L}}\sum\limits_{(i',j')\in \mathbb{L'}}{\rm max}\{0,\tilde{d}(i,j)-\tilde{d}(i',j')+\gamma\}
\end{equation}

where $\gamma > 0$ is a margin hyper-parameter; $\mathbb{L}$ is our alignment seeds and $\mathbb{L'}$ is the set of negative aligned entity pairs generated by nearest neighbor
sampling \citep{kotnis2017analysis}. 
\paragraph{Overall training objective.} The pre-training phase terminates once the entity alignment performance has converged to be stable. We find that after this stage, the entity representations given by the GCN are sufficient for supporting the neighborhood sampling and matching modules. Hence, we replace the loss function of NMN after the pre-training phase as: 
\begin{equation}
\label{overall_training_obj}
\small
L=\sum\limits_{(r,t)\in \mathbb{L}}\sum\limits_{(r',t')\in \mathbb{C}}{\rm max}\{0,d(r,t)-d(r',t')+\gamma\}
\end{equation}
\vspace{-3mm}
\begin{equation}
\small
d(r,t)=\|\bm{h}^{match}_r-\bm{h}^{match}_t\|_{L_1}
\end{equation}
where the negative alignments set $\mathbb{C}=\{(r',t')|(r'=r\wedge t'\in \mathcal{C}_r)\vee(t'=t\wedge r'\in \mathcal{C}_t)\}$ is made up of the alignment candidate sets of $r$ and $t$, $\mathcal{C}_r$ and $\mathcal{C}_t$ are generated in the candidate selection stage described in Sec. \ref{neighbor_match}.

\begin{figure}[t!]
    \centering
	\includegraphics[width=0.7\linewidth]{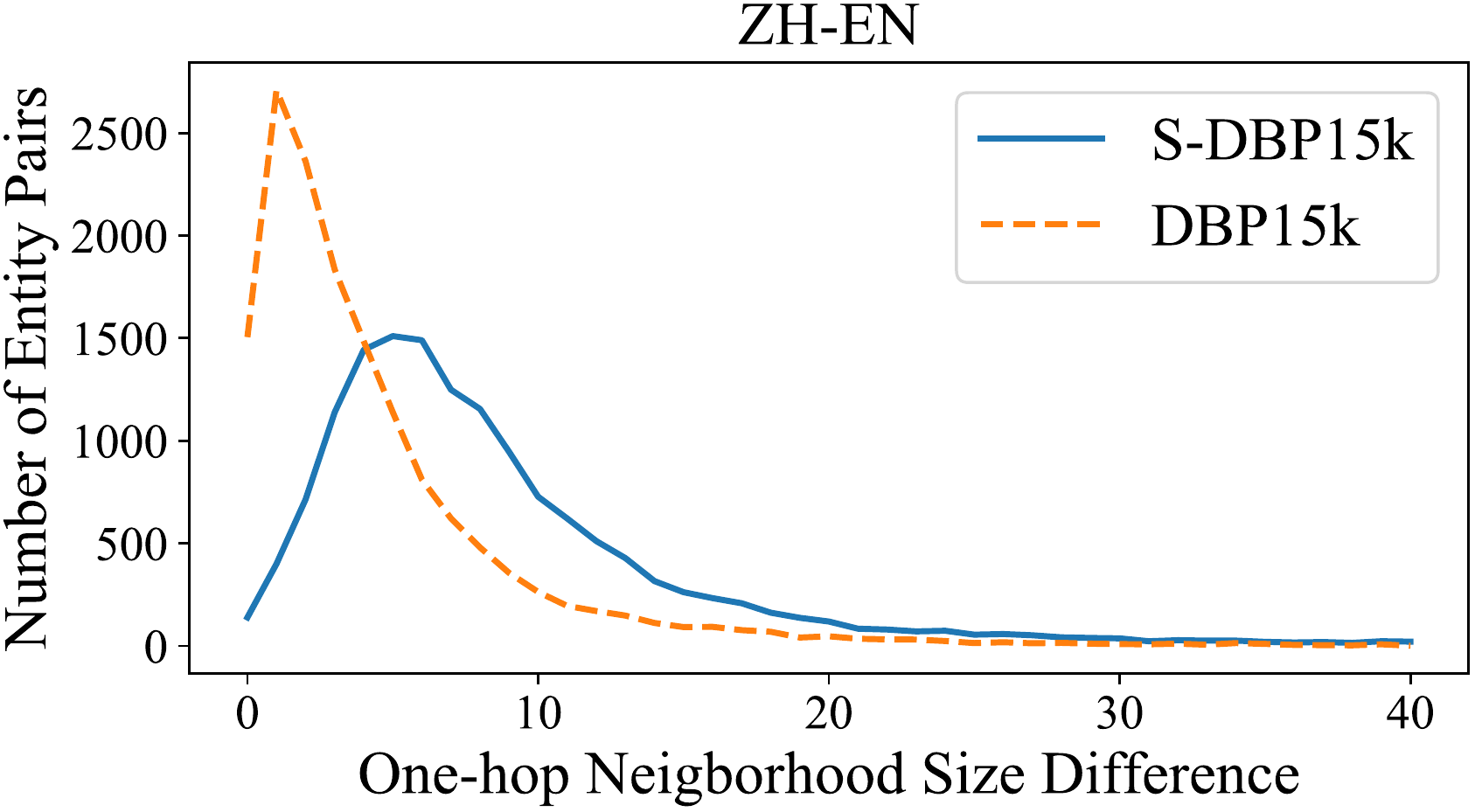}
	\vspace{-2mm}
	\caption{Distribution of difference in the size of neighborhoods of aligned entity pairs on DBP15K$_{ZH-EN}$.}
	\label{FigDiff}
	\vspace{-4mm}
\end{figure}

Note that our sampling process is non-differentiable, which corrupts the training of weight matrix $\textbf{W}_s$ in Eq. \ref{sample_distribution}. To avoid this issue, when training $\textbf{W}_s$, instead of direct sampling, we aggregate all the neighbor information by intuitive weighted summation:
\begin{equation}
\bm{g}^w_i=(\sum_{p\in N_i}\alpha_{ip}\cdot \sigma(\hat{\bm{h}}_p \textbf{W}_{gate})\cdot \hat{\bm{h}}_p)\textbf{W}_N
\vspace{-2mm}
\end{equation}
where $\alpha_{ip}$ is the aggregation weight for neighbor $p$, and is the sampling probability $p(\bm{h}_{p}|\bm{h}_i)$ for $p$ given by Eq.~\ref{sample_distribution}. Since the aim of training $\textbf{W}_s$ is to let the learned neighborhood representations of aligned entities to be as similar as possible, the objective is:
\begin{equation}
\label{training_W}
L_w=\sum\limits_{(r,t)\in \mathbb{L}}\|\bm{g}^w_r-\bm{g}^w_t\|_{L_1}
\end{equation}

In general, our model is trained end-to-end after pre-training. During training, we use Eq. \ref{overall_training_obj} as the main objective function, and, every 50 epochs, we tune $\textbf{W}_s$ using Eq. \ref{training_W} as the objective function.

\section{Experimental Setup}

\paragraph{Datasets.}
Follow the common practice of recent works~\citep{ijcai2018-611,cao2019multi,sun2019knowledge}, we evaluate our model on DBP15K \citep{sun2017cross} and DWY100K \citep{ijcai2018-611} datasets, and use the same split with previous works, 30\% for training and 70\% for testing.
To evaluate the performance of NMN in a more challenging setting, we also build a sparse dataset S-DBP15K based on DBP15K. Specifically, we randomly remove a
certain proportion of triples in the non-English KG to increase the difference in neighborhood size for entities in different KGs.
Table \ref{dataset} gives the detailed statistics of DBP15K and S-DBP15K, and the information of DWY100K is exhibited in Table \ref{dataset2}. Figure~\ref{FigDiff} shows the distribution of difference in
the size of one-hop neighborhoods of aligned entity pairs. Our source code and datasets are freely available online.\footnote{https://github.com/StephanieWyt/NMN}

\begin{table}[t!]
	\centering
	\scriptsize
	\begin{tabular}{l|l|ccccc}
		\toprule
		\multicolumn{2}{c|}{\textbf{Datasets}} & \textbf{Ent.} & \textbf{Rel.} & \textbf{Tri.} & \textbf{Tri. Remain in S.} \\
		\midrule
		\multirow{2}{*}{ZH-EN} & ZH & 66,469 & 2,830 & 153,929 & 26\% \\
		& EN & 98,125 & 2,317 & 237,674 & 100\% \\
		\midrule
		\multirow{2}{*}{JA-EN} & JA & 65,744 & 2,043 & 164,373 & 41\% \\
		& EN & 95,680 & 2,096 & 233,319 & 100\% \\
		\midrule
		\multirow{2}{*}{FR-EN} & FR & 66,858 & 1,379 & 192,191 & 45\% \\
		& EN & 105,889 & 2,209 & 278,590 & 100\% \\
		\bottomrule
	\end{tabular}
	\vspace{-2mm}
	\caption{Summary of DBP15K and S-DBP15k.}
	\label{dataset}
\end{table}

\begin{table}[t!]
	\centering
	\scriptsize
	\begin{tabular}{l|l|cccc}
		\toprule
		\multicolumn{2}{c|}{\textbf{Datasets}} & \textbf{Ent.} & \textbf{Rel.} & \textbf{Tri.} \\
		\midrule
		\multirow{2}{*}{DBP-WD} & DBpedia & 100,000 & 330 & 463,294 \\
		& Wikidata & 100,000 & 220 & 448,774 \\
		\midrule
		\multirow{2}{*}{DBP-YG} & DBpedia & 100,000 & 302 & 428,952 \\
		& YAGO3 & 100,000 & 31 & 502,563 \\
		\bottomrule
	\end{tabular}
	\vspace{-2mm}
	\caption{Summary of DWY100K.}
	\label{dataset2}
	\vspace{-3mm}
\end{table}

\paragraph{Comparison models.}
We compare NMN against 12 recently proposed embedding-based alignment methods: MTransE \citep{chen2016multilingual},
JAPE \citep{sun2017cross}, IPTransE \citep{zhu2017iterative}, GCN-Align \citep{D18-1032}, BootEA \citep{ijcai2018-611},
SEA \citep{pei2019semi}, RSN \citep{guo2019learning}, MuGNN \citep{cao2019multi}, KECG \citep{li2019semi}, AliNet
\citep{sun2019knowledge}, GMNN \citep{gmnn_acl19} and RDGCN \citep{Wu:2019:REA:3367722.3367779}. The last two models
also utilize entity names for alignment.

\paragraph{Model variants.}
To evaluate different components of our model, we provide two implementation variants of NMN: 
(1) NMN (w/o nbr-m), where we replace the neighborhood matching part by taking the average of sampled neighbor representations as the neighborhood representation; and
(2) NMN (w/o nbr-s), where we remove the sampling process and perform neighborhood matching on all one-hop neighbors.

\paragraph{Implementation details.}

The configuration we use in the DBP15K and DWY100k datasets is: $\beta=0.1$, $\gamma=1.0$, and we sample 5 neighbors for each entity in the neighborhood
sampling stage (Sec.~\ref{neighbor_sample}). For S-DBP15K, we set $\beta$ to 1. We sample 3 neighbors for each entity in S-DBP15K$_{ZH-EN}$ and S-DBP15K$_{JA-EN}$, and 10 neighbors in S-DBP15K$_{FR-EN}$. NMN uses a 2-layer GCN. The dimension of
hidden representations in GCN layers described in Sec.~\ref{structure-embed} is 300, and the dimension of neighborhood representation
$\bm{g}_i$ described in Sec.~\ref{neighborhood-agg} is 50. The size of the candidate set in Sec.~\ref{neighbor_match} is 20 for each
entity. The learning rate is set to 0.001. 

To initialize entity names, for the DBP15K datasets, we first use
Google Translate to translate all non-English entity names into English, and use pre-trained English word vectors
\emph{glove.840B.300d}\footnote{http://nlp.stanford.edu/projects/glove/} to construct the initial node features of KGs. For the DWY100K datasets, we directly use the pre-trained word vectors to initialize the nodes.

\paragraph{Metrics.} Following convention, we use Hits@1 and Hits@10 as our evaluation metrics. A Hits@k score is computed
by measuring the proportion of correctly aligned entities ranked in the top $k$ list. A higher Hits@k score indicates better performance.

\section{Experimental Results}
\begin{table*}[t!]
	\centering
	\scriptsize
	\begin{tabular}{p{90pt}|p{18pt}<{\centering}p{18pt}<{\centering}|p{18pt}<{\centering}p{18pt}<{\centering}|p{18pt}<{\centering}p{18pt}<{\centering}|p{18pt}<{\centering}p{18pt}<{\centering}|p{18pt}<{\centering}p{18pt}<{\centering}}
		\toprule
		\multirow{2}{*}{\bf Models} & \multicolumn{2}{c|}{\bf $\textbf{DBP}_\textbf{ZH-EN}$} & \multicolumn{2}{c|}{\bf $\textbf{DBP}_\textbf{JA-EN}$} & \multicolumn{2}{c|}{ \bf $\textbf{DBP}_\textbf{FR-EN}$} &
		\multicolumn{2}{c|}{\bf DBP-WD} & \multicolumn{2}{c}{ \bf DBP-YG} \\
		& \tiny \bf Hits@1 & \tiny \bf Hits@10 & \tiny \bf Hits@1 &\tiny \bf Hits@10 & \tiny \bf Hits@1 &\tiny \bf Hits@10 & \tiny \bf Hits@1 & \tiny \bf Hits@10 & \tiny \bf Hits@1 & \tiny \bf Hits@10 \\
		\midrule
		MTransE \citep{chen2016multilingual} & 30.8 & 61.4 & 27.9 & 57.5 & 24.4 & 55.6 & 28.1 & 52.0 & 25.2 & 49.3\\
		JAPE \citep{sun2017cross} & 41.2 & 74.5 & 36.3 & 68.5 & 32.4 & 66.7 & 31.8 & 58.9 & 23.6 & 48.4 \\
		IPTransE \citep{zhu2017iterative} & 40.6 & 73.5 & 36.7 & 69.3 & 33.3 & 68.5 & 34.9 & 63.8 & 29.7 & 55.8 \\
		GCN-Align \citep{D18-1032} & 41.3 & 74.4 & 39.9 & 74.5 & 37.3 & 74.5 & 50.6 & 77.2 & 59.7 & 83.8 \\
		SEA \citep{pei2019semi} & 42.4 & 79.6 & 38.5 & 78.3 & 40.0 & 79.7 & 51.8 & 80.2 & 51.6 & 73.6 \\
		RSN \citep{guo2019learning} & 50.8 & 74.5 & 50.7 & 73.7 & 51.6 & 76.8 & 60.7 & 79.3 & 68.9 & 87.8 \\
		KECG \citep{li2019semi} & 47.8 & 83.5 & 49.0 & 84.4 & 48.6 & 85.1 & 63.2 & 90.0 & 72.8 & 91.5 \\
        MuGNN \citep{cao2019multi} & 49.4 & 84.4 & 50.1 & 85.7 & 49.5 & 87.0 & 61.6 & 89.7 & 74.1 & 93.7 \\
		AliNet \citep{sun2019knowledge} & 53.9 & 82.6 & 54.9 & 83.1 & 55.2 & 85.2 & 69.0 & 90.8 & 78.6 & 94.3\\
		BootEA \citep{ijcai2018-611} & 62.9 & 84.8 & 62.2 & 85.4 & 65.3 & 87.4 & 74.8 & 89.8 & 76.1 & 89.4 \\
		\midrule
		GMNN \citep{gmnn_acl19} & 67.9 & 78.5 & 74.0 & 87.2 & 89.4 & 95.2 & 93.0 & \bf 99.6 & 94.4 & \bf 99.8 \\
		RDGCN \citep{Wu:2019:REA:3367722.3367779} & 70.8 & 84.6 & 76.7 & 89.5 & 88.6 & 95.7 & 97.9 & 99.1 & 94.7 & 97.3 \\
		\midrule
		\rowcolor{Gray} \bf NMN & \bf 73.3 & \bf 86.9 & \bf 78.5 & \bf 91.2 & \bf 90.2 & \bf 96.7 & \bf 98.1 & 99.2 & \bf 96.0 & 98.2 \\
		\quad w/o nbr-m & 71.1 & 86.7 & 75.4 & 90.4 & 86.3 & 95.8 & 96.0 & 98.4 & 95.0 & 97.8 \\
		\quad w/o nbr-s & 73.0 & 85.6 & 77.9 & 88.8 & 89.9 & 95.7 & 98.0 & 99.0 & 95.9 & 98.1 \\
		\bottomrule
	\end{tabular}
	\vspace{-1mm}
	\caption{Performance on DBP15K and DWY100K.}
	\label{Fullresults}
	\vspace{-2mm}
\end{table*}

\subsection{Performance on DBP15K and DWY100K} Table \ref{Fullresults} reports the entity alignment performance of all approaches on DBP15K and DWY100K datasets. It shows that the full implementation of NMN significantly outperforms all alternative approaches.

\paragraph{Structured-based methods.}
The top part of the table shows the performance of the state-of-the-art structure-based models which solely utilize structural
information. Among them, BootEA delivers the best performance where it benefits from more training instances through a bootstrapping
process. By considering the structural heterogeneity, MuGNN and AliNet outperform most of other structure-based counterparts, showing the
importance of tackling structural heterogeneity.

\paragraph{Entity name initialization.}
The middle part of Table \ref{Fullresults} gives the results of embedding-based models that use entity name information
along with structural information. Using entity names to initialize node features, the GNN-based models, GMNN and RDGCN,
show a clear improvement over structure-based models, suggesting that entity names provide useful clues for entity
alignment. In particular, GMNN achieves the highest Hits@10 on the DWY100K datasets, which are the only monolingual datasets (in English) in our experiments. 
We also note that, GMNN pre-screens a small candidate set for each entity based on the entity name similarity, and only traverses  this candidate set during testing and calculating the Hits@k scores.

\paragraph{NMN vs. its variants.}
The bottom part of Table~\ref{Fullresults} shows the performance of NMN and its variants. Our full NMN implementation substantially
outperforms all baselines across nearly all metrics and datasets by accurately modeling entity neighborhoods through neighborhood sampling and
matching and using entity name information. Specifically, NMN achieves the best Hits@1 score on DBP15K$_{ZH-EN}$, with a gain of 2.5\% compared with RDGCN, and
5.4\% over GMNN. Although RDGCN employs a dual relation graph to model the complex relation information, it does not address the issue of
neighborhood heterogeneity. While GMNN collects all one-hop neighbors to construct a topic entity graph for each entity, its strategy might
introduce noises since not all one-hop neighbors are favorable for entity alignment.

When comparing NMN and NMN (w/o nbr-m), we can observe around a 2.5\% drop in Hits@1 and a 0.6\% drop in Hits@10 on average, after removing the neighborhood matching module. Specifically, the Hits@1 scores
between NMN and NMN (w/o nbr-m) differ by 3.9\% on DBP15K$_{FR-EN}$. 
These results confirm the effectiveness of our neighborhood matching module in identifying matching neighbors and estimating the neighborhood similarity.

Removing the neighbor sampling module from NMN, i.e., NMN (w/o nbr-s), leads to an average performance drop of 0.3\% on Hits@1 and 1\% on Hits@10 on all the datasets. This result shows the important role of our sampling module in filtering irrelevant neighbors.

When removing either the neighborhood matching module (NMN (w/o nbr-m)) or sampling module (NMN (w/o nbr-s)) from our main model, we see a substantially larger drop in both Hits@1 and Hits@10 on DBP15K than on DWY100K. One reason is that the heterogeneity problem in DBP15K is more severe than that in DWY100K. The average proportion of aligned entity pairs that have a different number of neighbors is 89\% in DBP15K compared to 84\% in DWY100K. These results show that our sampling and matching modules are particularly important, when the neighborhood sizes of equivalent entities greatly differ and especially there may be few common neighbors in their neighborhoods.

\begin{table}[t!]
	\centering
	\scriptsize
	\begin{tabular}{p{38pt}|p{14pt}<{\centering}p{18pt}<{\centering}|p{14pt}<{\centering}p{18pt}<{\centering}|p{14pt}<{\centering}p{18pt}<{\centering}}
		\toprule
		\multirow{2}{*}{\bf Models} & \multicolumn{2}{c|}{\bf ZH-EN} & \multicolumn{2}{c|}{\bf JA-EN} & \multicolumn{2}{c}{ \bf FR-EN}  \\
		& \tiny \bf Hits@1 & \tiny \bf Hits@10 & \tiny \bf Hits@1 &\tiny \bf Hits@10 & \tiny \bf Hits@1 &\tiny \bf Hits@10\\
		\midrule
		BootEA & 12.2 & 27.5 & 27.8 & 52.6 & 32.7 & 53.2 \\
		\midrule
		GMNN & 47.5 & 68.3 & 58.8 & 78.2 & 75.0 & 90.9 \\
		RDGCN & 60.7 & 74.6 & 69.3 & 82.9 & 83.6 & 92.6 \\
		\midrule
		\rowcolor{Gray} \bf NMN & \bf 62.0 & \bf 75.1 & 70.3 & 84.4 & 86.3 & 94.0 \\
		\quad w/o nbr-m & 52.0 & 71.1 & 62.1 & 82.7 & 80.0 & 92.0 \\
		\quad w/o nbr-s & 60.9 & 74.1 & \bf 70.7 & \bf 84.5 & \bf 86.5 & \bf 94.2 \\
		\bottomrule
	\end{tabular}
	\vspace{-1mm}
	\caption{Performance on S-DBP15K.}
	\label{Newresults}
	\vspace{-3mm}
\end{table}

\subsection{Performance on S-DBP15K}
On the more sparse and challenging datasets S-DBP15K, we compare our NMN model with the strongest
structure-based model, BootEA, and GNN-based models, GMNN and RDGCN, which also utilize the entity name initialization.

\paragraph{Baseline models.}
In Table~\ref{Newresults}, we can observe that all models suffer a performance drop, where BootEA endures the most significant drop.
With the support of entity names, GMNN and RDGCN achieve better performances over BootEA. These results show when the alignment clues are
sparse, structural information alone is not sufficient to support precise comparisons, and the entity name semantics are particularly useful for accurate alignment in such case.

\paragraph{NMN.}
Our NMN outperforms all three baselines on all sparse datasets,
demonstrating the effectiveness and robustness of NMN. As discussed in Sec.~\ref{section:intro}, the performances of existing embedding-based
methods decrease significantly as the gap of equivalent entities' neighborhood sizes increases. Specifically, on DBP15K$_{ZH-EN}$, our NMN outperforms RDGCN,
the best-performing baseline, by a large margin, achieving 65\%, 53\% and 48\% on Hits@1 on the entity pairs whose number of neighbors
differs by more than 10, 20 and 30, respectively.

\begin{figure*}[t!]
    \centering
	\includegraphics[width=0.9\linewidth]{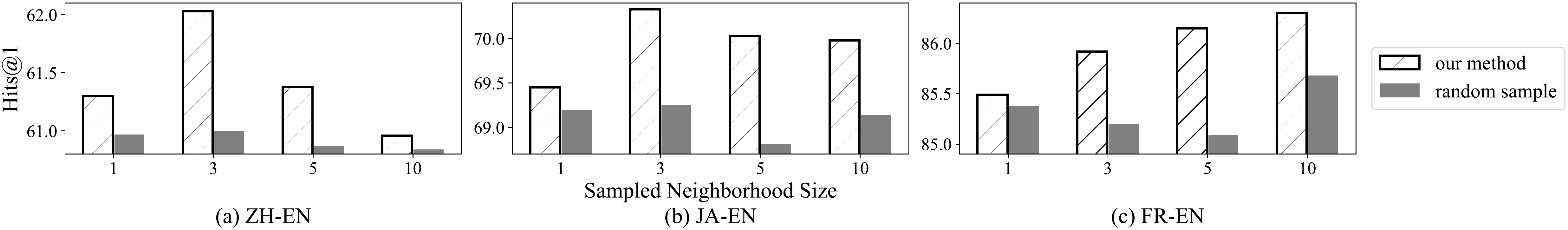}
	\vspace{-2mm}
	\caption{Comparison between our neighborhood sampling strategy and random sampling on S-DBP15K.}
	\label{FigSample}
	\vspace{-3mm}
\end{figure*}

\begin{figure}[t!]
    \centering
	\includegraphics[width=0.8\linewidth, trim=0 30 0 10,clip]{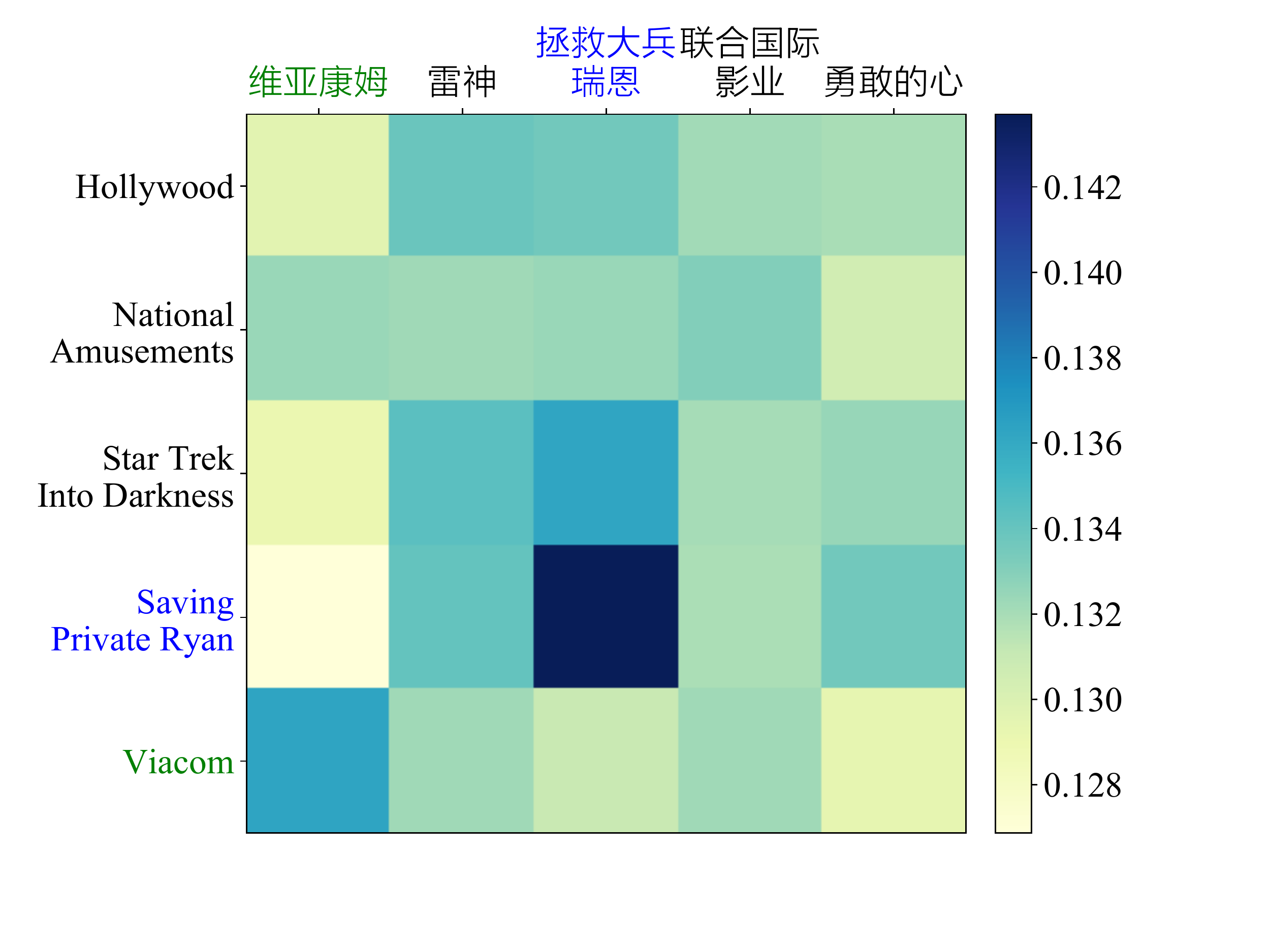}
	\vspace{-3mm}
	\caption{Visualization of attention weights in the neighborhood matching module for the example of \emph{Paramount Pictures}. The green and blue words are two pairs of equivalent neighbors.}
	\vspace{-4mm}
	\label{FigWeight}
\end{figure}

\paragraph{Sampling and matching strategies.}
When we compare NMN and NMN (w/o nbr-m) on the S-DBP15K, we can see a larger average drop in Hits@1 than on the DBP15K (8.2\% vs. 3.1\%). The result indicates that our neighborhood matching module plays a more important role on the more sparse dataset. When the alignment clues are less obvious, our matching module can continuously amplify the neighborhood difference of an entity pair during the propagation
process. In this way, the gap between the equivalent entity pair and the negative pairs becomes larger, leading to correct alignment.

Compared with NMN, removing sampling module does hurt NMN in both Hits@1 and Hits@10 on S-DBP15K$_{ZH-EN}$. But, it is surprising that NMN (w/o nbr-s) delivers slightly better results than NMN on S-DBP15K$_{JA-EN}$ and S-DBP15K$_{FR-EN}$. Since the average number of neighbors of entities in S-DBP15K is much less than that in the DBP15K datasets. When the number of neighbors is small, the role of sampling will be unstable. In addition, our sampling method is relatively simple. When the alignment clues are very sparse, our strategy may not be robust enough. We will explore more adaptive sampling method and scope in the future.

\subsection{Analysis}
\label{analyses}
\paragraph{Impact of neighborhood sampling strategies.} To explore the impact of neighborhood sampling strategies, we compare our
NMN with a variant that uses random sampling strategy on S-DBP15K datasets. Figure \ref{FigSample} illustrates
the Hits@1 of NMN using our designed graph sampling method (Sec.~\ref{neighbor_sample}) and a random-sampling-based
variant when sampling different number of neighbors. Our NMN consistently delivers better results compared to the variant, showing that our sampling strategy can effectively select more informative neighbors. 

\paragraph{Impact of neighborhood sampling size.} From Figure \ref{FigSample}, for S-DBP15K$_{ZH-EN}$, both models reach a performance plateau with a sampling size of 3, and using a bigger sampling size would lead to performance degradation. For S-DBP15K$_{JA-EN}$ and S-DBP15K$_{FR-EN}$, we observe that our NMN performs similarly when sampling different number of neighbors. From Table \ref{dataset}, we can see that S-DBP15K$_{ZH-EN}$ is more sparse than S-DBP15K$_{JA-EN}$ and S-DBP15K$_{FR-EN}$. All models deliver much lower performance on S-DBP15K$_{ZH-EN}$. Therefore, the neighbor quality of this dataset might be poor, and a larger sampling size will introduce more noise. On the other hand, the neighbors in JA-EN and FR-EN datasets might be more informative. Thus, NMN is not sensitive to the sampling size on these two datasets.

\paragraph{How does the neighborhood matching module work?}
In an attempt to understand how our neighborhood matching strategy helps alignment, we visualize the attention weights in the neighborhood
matching module. Considering an equivalent entity pair in DBP15K$_{ZH-EN}$, both
of which indicate an American film studio \emph{Paramount Pictures}. 
From Figure~\ref{FigWeight}, we can see that the five neighbors sampled by our sampling module for each central entity are very informative ones for aligning the two central entities, such as the famous movies released by \emph{Paramount Pictures}, the parent company and subsidiary of \emph{Paramount Pictures}. This demonstrates the effectiveness of our sampling strategy again. Among the sampled neighbors, there are also two pairs of common neighbors (indicate \emph{Saving Private Ryan} and \emph{Viacom}). We observe that for each pair of equivalent neighbors, one neighbor can be particularly attended by its counterpart (the corresponding square has a darker color). This example clearly demonstrates that our neighborhood matching module can accurately estimate the neighborhood similarity by accurately detecting the similar neighbors.

\section{Conclusion}
We have presented NMN, a novel embedded-based framework for entity alignment. NMN tackles the ubiquitous neighborhood
heterogeneity in KGs. We achieve this by using a new sampling-based approach to choose the most informative neighbors for each entity. As a
departure from prior works, NMN simultaneously estimates the similarity of two entities, by considering both topological structure and
neighborhood similarity. We perform extensive experiments on real-world datasets and compare NMN against 12 recent embedded-based methods. 
Experimental results show that NMN achieves the best and more robust performance, 
 consistently outperforming competitive methods across datasets and evaluation metrics. 

\section*{Acknowledgments}
This work is supported in part by the National Hi-Tech R\&D Program of China (No. 2018YFB1005100), the NSFC under grant agreements 61672057, 61672058 and 61872294, and a UK Royal Society International Collaboration Grant. For any correspondence, please contact Yansong Feng.

\bibliography{acl2020}

\begin{thebibliography}{35}
\expandafter\ifx\csname natexlab\endcsname\relax\def\natexlab#1{#1}\fi

\bibitem[{Bastings et~al.(2017)Bastings, Titov, Aziz, Marcheggiani, and
  Sima{'}an}]{Bastings2017Graph}
Joost Bastings, Ivan Titov, Wilker Aziz, Diego Marcheggiani, and Khalil
  Sima{'}an. 2017.
\newblock \href {https://doi.org/10.18653/v1/D17-1209} {Graph convolutional
  encoders for syntax-aware neural machine translation}.
\newblock In \emph{Proceedings of the 2017 Conference on Empirical Methods in
  Natural Language Processing}, pages 1957--1967, Copenhagen, Denmark.
  Association for Computational Linguistics.

\bibitem[{Bordes et~al.(2013)Bordes, Usunier, Garcia-Duran, Weston, and
  Yakhnenko}]{bordes2013translating}
Antoine Bordes, Nicolas Usunier, Alberto Garcia-Duran, Jason Weston, and Oksana
  Yakhnenko. 2013.
\newblock \href
  {http://papers.nips.cc/paper/5071-translating-embeddings-for-modeling-multi-relational-data.pdf}
  {Translating embeddings for modeling multi-relational data}.
\newblock In \emph{Advances in Neural Information Processing Systems 26}, pages
  2787--2795. Curran Associates, Inc.

\bibitem[{Cao et~al.(2019)Cao, Liu, Li, Liu, Li, and Chua}]{cao2019multi}
Yixin Cao, Zhiyuan Liu, Chengjiang Li, Zhiyuan Liu, Juanzi Li, and Tat-Seng
  Chua. 2019.
\newblock \href {https://doi.org/10.18653/v1/P19-1140} {Multi-channel graph
  neural network for entity alignment}.
\newblock In \emph{Proceedings of the 57th Annual Meeting of the Association
  for Computational Linguistics}, pages 1452--1461, Florence, Italy.
  Association for Computational Linguistics.

\bibitem[{Chen et~al.(2018)Chen, Tian, Chang, Skiena, and Zaniolo}]{Chen2018Co}
Muhao Chen, Yingtao Tian, Kai{-}Wei Chang, Steven Skiena, and Carlo Zaniolo.
  2018.
\newblock \href {https://doi.org/10.24963/ijcai.2018/556} {Co-training
  embeddings of knowledge graphs and entity descriptions for cross-lingual
  entity alignment}.
\newblock In \emph{Proceedings of the Twenty-Seventh International Joint
  Conference on Artificial Intelligence, {IJCAI} 2018, July 13-19, 2018,
  Stockholm, Sweden}, pages 3998--4004. ijcai.org.

\bibitem[{Chen et~al.(2017)Chen, Tian, Yang, and
  Zaniolo}]{chen2016multilingual}
Muhao Chen, Yingtao Tian, Mohan Yang, and Carlo Zaniolo. 2017.
\newblock \href {https://doi.org/10.24963/ijcai.2017/209} {Multilingual
  knowledge graph embeddings for cross-lingual knowledge alignment}.
\newblock In \emph{Proceedings of the Twenty-Sixth International Joint
  Conference on Artificial Intelligence, {IJCAI} 2017, Melbourne, Australia,
  August 19-25, 2017}, pages 1511--1517. ijcai.org.

\bibitem[{Guo et~al.(2019)Guo, Sun, and Hu}]{guo2019learning}
Lingbing Guo, Zequn Sun, and Wei Hu. 2019.
\newblock \href {http://proceedings.mlr.press/v97/guo19c.html} {Learning to
  exploit long-term relational dependencies in knowledge graphs}.
\newblock In \emph{Proceedings of the 36th International Conference on Machine
  Learning, {ICML} 2019, 9-15 June 2019, Long Beach, California, {USA}},
  volume~97 of \emph{Proceedings of Machine Learning Research}, pages
  2505--2514. {PMLR}.

\bibitem[{Hu and Lau(2013)}]{hu2013survey}
Pili Hu and Wing~Cheong Lau. 2013.
\newblock \href {http://arxiv.org/abs/1308.5865} {A survey and taxonomy of
  graph sampling}.
\newblock \emph{arXiv preprint arXiv:1308.5865}.

\bibitem[{Kipf and Welling(2017)}]{Kipf2016Semi}
Thomas~N. Kipf and Max Welling. 2017.
\newblock \href {https://openreview.net/forum?id=SJU4ayYgl} {Semi-supervised
  classification with graph convolutional networks}.
\newblock In \emph{5th International Conference on Learning Representations,
  {ICLR} 2017, Toulon, France, April 24-26, 2017, Conference Track
  Proceedings}. OpenReview.net.

\bibitem[{Kotnis and Nastase(2017)}]{kotnis2017analysis}
Bhushan Kotnis and Vivi Nastase. 2017.
\newblock \href {http://arxiv.org/abs/1708.06816} {Analysis of the impact of
  negative sampling on link prediction in knowledge graphs}.
\newblock \emph{arXiv preprint arXiv:1708.06816}.

\bibitem[{Li et~al.(2019{\natexlab{a}})Li, Cao, Hou, Shi, Li, and
  Chua}]{li2019semi}
Chengjiang Li, Yixin Cao, Lei Hou, Jiaxin Shi, Juanzi Li, and Tat-Seng Chua.
  2019{\natexlab{a}}.
\newblock \href {https://doi.org/10.18653/v1/D19-1274} {Semi-supervised entity
  alignment via joint knowledge embedding model and cross-graph model}.
\newblock In \emph{Proceedings of the 2019 Conference on Empirical Methods in
  Natural Language Processing and the 9th International Joint Conference on
  Natural Language Processing (EMNLP-IJCNLP)}, pages 2723--2732, Hong Kong,
  China. Association for Computational Linguistics.

\bibitem[{Li et~al.(2019{\natexlab{b}})Li, Gu, Dullien, Vinyals, and
  Kohli}]{li2019graph}
Yujia Li, Chenjie Gu, Thomas Dullien, Oriol Vinyals, and Pushmeet Kohli.
  2019{\natexlab{b}}.
\newblock \href {http://proceedings.mlr.press/v97/li19d.html} {Graph matching
  networks for learning the similarity of graph structured objects}.
\newblock In \emph{Proceedings of the 36th International Conference on Machine
  Learning, {ICML} 2019, 9-15 June 2019, Long Beach, California, {USA}},
  volume~97 of \emph{Proceedings of Machine Learning Research}, pages
  3835--3845. {PMLR}.

\bibitem[{Li et~al.(2016)Li, Zemel, Brockschmidt, and Tarlow}]{li2015gated}
Yujia Li, Richard Zemel, Marc Brockschmidt, and Daniel Tarlow. 2016.
\newblock \href {http://arxiv.org/abs/1511.05493} {Gated graph sequence neural
  networks}.
\newblock In \emph{Proceedings of ICLR'16}.

\bibitem[{Mahdisoltani et~al.(2015)Mahdisoltani, Biega, and
  Suchanek}]{Mahdisoltani2014YAGO3AK}
Farzaneh Mahdisoltani, Joanna Biega, and Fabian~M. Suchanek. 2015.
\newblock \href {http://cidrdb.org/cidr2015/Papers/CIDR15\_Paper1.pdf}
  {{YAGO3:} {A} knowledge base from multilingual wikipedias}.
\newblock In \emph{{CIDR} 2015, Seventh Biennial Conference on Innovative Data
  Systems Research, Asilomar, CA, USA, January 4-7, 2015, Online Proceedings}.
  www.cidrdb.org.

\bibitem[{Marcheggiani and Titov(2017)}]{Marcheggiani2017Encoding}
Diego Marcheggiani and Ivan Titov. 2017.
\newblock \href {https://doi.org/10.18653/v1/D17-1159} {Encoding sentences with
  graph convolutional networks for semantic role labeling}.
\newblock In \emph{Proceedings of the 2017 Conference on Empirical Methods in
  Natural Language Processing}, pages 1506--1515, Copenhagen, Denmark.
  Association for Computational Linguistics.

\bibitem[{Pei et~al.(2019{\natexlab{a}})Pei, Yu, Hoehndorf, and
  Zhang}]{pei2019semi}
Shichao Pei, Lu~Yu, Robert Hoehndorf, and Xiangliang Zhang. 2019{\natexlab{a}}.
\newblock \href {https://doi.org/10.1145/3308558.3313646} {Semi-supervised
  entity alignment via knowledge graph embedding with awareness of degree
  difference}.
\newblock In \emph{The World Wide Web Conference, {WWW} 2019, San Francisco,
  CA, USA, May 13-17, 2019}, pages 3130--3136. {ACM}.

\bibitem[{Pei et~al.(2019{\natexlab{b}})Pei, Yu, and Zhang}]{pei2019improving}
Shichao Pei, Lu~Yu, and Xiangliang Zhang. 2019{\natexlab{b}}.
\newblock \href {https://doi.org/10.24963/ijcai.2019/448} {Improving
  cross-lingual entity alignment via optimal transport}.
\newblock In \emph{Proceedings of the Twenty-Eighth International Joint
  Conference on Artificial Intelligence, {IJCAI} 2019, Macao, China, August
  10-16, 2019}, pages 3231--3237. ijcai.org.

\bibitem[{Rahimi et~al.(2018)Rahimi, Cohn, and Baldwin}]{Rahimi2018Semi}
Afshin Rahimi, Trevor Cohn, and Timothy Baldwin. 2018.
\newblock \href {https://doi.org/10.18653/v1/P18-1187} {Semi-supervised user
  geolocation via graph convolutional networks}.
\newblock In \emph{Proceedings of the 56th Annual Meeting of the Association
  for Computational Linguistics (Volume 1: Long Papers)}, pages 2009--2019,
  Melbourne, Australia. Association for Computational Linguistics.

\bibitem[{Raymond et~al.(2002)Raymond, Gardiner, and
  Willett}]{raymond2002rascal}
John~W. Raymond, Eleanor~J. Gardiner, and Peter Willett. 2002.
\newblock \href {https://doi.org/10.1093/comjnl/45.6.631} {{RASCAL:}
  calculation of graph similarity using maximum common edge subgraphs}.
\newblock \emph{Comput. J.}, 45(6):631--644.

\bibitem[{Schlichtkrull et~al.(2018)Schlichtkrull, Kipf, Bloem, van~den Berg,
  Titov, and Welling}]{schlichtkrull2018modeling}
Michael~Sejr Schlichtkrull, Thomas~N. Kipf, Peter Bloem, Rianne van~den Berg,
  Ivan Titov, and Max Welling. 2018.
\newblock \href {https://doi.org/10.1007/978-3-319-93417-4\_38} {Modeling
  relational data with graph convolutional networks}.
\newblock In \emph{The Semantic Web - 15th International Conference, {ESWC}
  2018, Heraklion, Crete, Greece, June 3-7, 2018, Proceedings}, volume 10843 of
  \emph{Lecture Notes in Computer Science}, pages 593--607. Springer.

\bibitem[{Srivastava et~al.(2015)Srivastava, Greff, and
  Schmidhuber}]{Srivastava2015Highway}
Rupesh~Kumar Srivastava, Klaus Greff, and J{\"u}rgen Schmidhuber. 2015.
\newblock \href {http://arxiv.org/abs/1505.00387} {Highway networks}.
\newblock \emph{arXiv preprint arXiv:1505.00387}.

\bibitem[{Sun et~al.(2017)Sun, Hu, and Li}]{sun2017cross}
Zequn Sun, Wei Hu, and Chengkai Li. 2017.
\newblock \href {https://doi.org/10.1007/978-3-319-68288-4\_37} {Cross-lingual
  entity alignment via joint attribute-preserving embedding}.
\newblock In \emph{The Semantic Web - {ISWC} 2017 - 16th International Semantic
  Web Conference, Vienna, Austria, October 21-25, 2017, Proceedings, Part {I}},
  volume 10587 of \emph{Lecture Notes in Computer Science}, pages 628--644.
  Springer.

\bibitem[{Sun et~al.(2018)Sun, Hu, Zhang, and Qu}]{ijcai2018-611}
Zequn Sun, Wei Hu, Qingheng Zhang, and Yuzhong Qu. 2018.
\newblock \href {https://doi.org/10.24963/ijcai.2018/611} {Bootstrapping entity
  alignment with knowledge graph embedding}.
\newblock In \emph{Proceedings of the Twenty-Seventh International Joint
  Conference on Artificial Intelligence, {IJCAI} 2018, July 13-19, 2018,
  Stockholm, Sweden}, pages 4396--4402. ijcai.org.

\bibitem[{Sun et~al.(2020)Sun, Wang, Hu, Chen, Dai, Zhang, and
  Qu}]{sun2019knowledge}
Zequn Sun, Chengming Wang, Wei Hu, Muhao Chen, Jian Dai, Wei Zhang, and Yuzhong
  Qu. 2020.
\newblock \href {https://aaai.org/Papers/AAAI/2020GB/AAAI-SunZ.3162.pdf}
  {Knowledge graph alignment network with gated multi-hop neighborhood
  aggregation}.
\newblock In \emph{AAAI}.

\bibitem[{Trisedya et~al.(2019)Trisedya, Qi, and Zhang}]{trisedya2019entity}
Bayu~Distiawan Trisedya, Jianzhong Qi, and Rui Zhang. 2019.
\newblock \href {https://doi.org/10.1609/aaai.v33i01.3301297} {Entity alignment
  between knowledge graphs using attribute embeddings}.
\newblock In \emph{Proceedings of the AAAI Conference on Artificial
  Intelligence}, volume~33, pages 297--304.

\bibitem[{Veli{\v{c}}kovi{\'{c}} et~al.(2018)Veli{\v{c}}kovi{\'{c}}, Cucurull,
  Casanova, Romero, Li{\`{o}}, and Bengio}]{velickovic2018graph}
Petar Veli{\v{c}}kovi{\'{c}}, Guillem Cucurull, Arantxa Casanova, Adriana
  Romero, Pietro Li{\`{o}}, and Yoshua Bengio. 2018.
\newblock \href {https://openreview.net/forum?id=rJXMpikCZ} {{Graph Attention
  Networks}}.
\newblock In \emph{ICLR}.

\bibitem[{Wang et~al.(2018)Wang, Lv, Lan, and Zhang}]{D18-1032}
Zhichun Wang, Qingsong Lv, Xiaohan Lan, and Yu~Zhang. 2018.
\newblock \href {https://doi.org/10.18653/v1/D18-1032} {Cross-lingual knowledge
  graph alignment via graph convolutional networks}.
\newblock In \emph{Proceedings of the 2018 Conference on Empirical Methods in
  Natural Language Processing}, pages 349--357, Brussels, Belgium. Association
  for Computational Linguistics.

\bibitem[{Wu et~al.(2019{\natexlab{a}})Wu, Liu, Feng, Wang, Yan, and
  Zhao}]{Wu:2019:REA:3367722.3367779}
Yuting Wu, Xiao Liu, Yansong Feng, Zheng Wang, Rui Yan, and Dongyan Zhao.
  2019{\natexlab{a}}.
\newblock \href {https://doi.org/10.24963/ijcai.2019/733} {Relation-aware
  entity alignment for heterogeneous knowledge graphs}.
\newblock In \emph{Proceedings of the Twenty-Eighth International Joint
  Conference on Artificial Intelligence, {IJCAI} 2019, Macao, China, August
  10-16, 2019}, pages 5278--5284. ijcai.org.

\bibitem[{Wu et~al.(2019{\natexlab{b}})Wu, Liu, Feng, Wang, and
  Zhao}]{wu2019jointly}
Yuting Wu, Xiao Liu, Yansong Feng, Zheng Wang, and Dongyan Zhao.
  2019{\natexlab{b}}.
\newblock \href {https://doi.org/10.18653/v1/D19-1023} {Jointly learning entity
  and relation representations for entity alignment}.
\newblock In \emph{Proceedings of the 2019 Conference on Empirical Methods in
  Natural Language Processing and the 9th International Joint Conference on
  Natural Language Processing (EMNLP-IJCNLP)}, pages 240--249, Hong Kong,
  China. Association for Computational Linguistics.

\bibitem[{Xu et~al.(2019)Xu, Wang, Yu, Feng, Song, Wang, and Yu}]{gmnn_acl19}
Kun Xu, Liwei Wang, Mo~Yu, Yansong Feng, Yan Song, Zhiguo Wang, and Dong Yu.
  2019.
\newblock \href {https://doi.org/10.18653/v1/P19-1304} {Cross-lingual knowledge
  graph alignment via graph matching neural network}.
\newblock In \emph{Proceedings of the 57th Annual Meeting of the Association
  for Computational Linguistics}, pages 3156--3161, Florence, Italy.
  Association for Computational Linguistics.

\bibitem[{Yan et~al.(2004)Yan, Yu, and Han}]{yan2004graph}
Xifeng Yan, Philip~S. Yu, and Jiawei Han. 2004.
\newblock \href {https://doi.org/10.1145/1007568.1007607} {Graph indexing: {A}
  frequent structure-based approach}.
\newblock In \emph{Proceedings of the {ACM} {SIGMOD} International Conference
  on Management of Data, Paris, France, June 13-18, 2004}, pages 335--346.
  {ACM}.

\bibitem[{Yang et~al.(2019)Yang, Zou, Shi, Lu, Lin, and Sun}]{yang2019aligning}
Hsiu-Wei Yang, Yanyan Zou, Peng Shi, Wei Lu, Jimmy Lin, and Xu~Sun. 2019.
\newblock \href {https://doi.org/10.18653/v1/D19-1451} {Aligning cross-lingual
  entities with multi-aspect information}.
\newblock In \emph{Proceedings of the 2019 Conference on Empirical Methods in
  Natural Language Processing and the 9th International Joint Conference on
  Natural Language Processing (EMNLP-IJCNLP)}, pages 4431--4441, Hong Kong,
  China. Association for Computational Linguistics.

\bibitem[{Ye et~al.(2019)Ye, Li, Fang, Zang, and Wang}]{ye2019vectorized}
Rui Ye, Xin Li, Yujie Fang, Hongyu Zang, and Mingzhong Wang. 2019.
\newblock \href {https://doi.org/10.24963/ijcai.2019/574} {A vectorized
  relational graph convolutional network for multi-relational network
  alignment}.
\newblock In \emph{Proceedings of the Twenty-Eighth International Joint
  Conference on Artificial Intelligence, {IJCAI} 2019, Macao, China, August
  10-16, 2019}, pages 4135--4141. ijcai.org.

\bibitem[{Zhang et~al.(2019)Zhang, Sun, Hu, Chen, Guo, and Qu}]{MultiKE}
Qingheng Zhang, Zequn Sun, Wei Hu, Muhao Chen, Lingbing Guo, and Yuzhong Qu.
  2019.
\newblock \href {https://doi.org/10.24963/ijcai.2019/754} {Multi-view knowledge
  graph embedding for entity alignment}.
\newblock In \emph{Proceedings of the Twenty-Eighth International Joint
  Conference on Artificial Intelligence, {IJCAI} 2019, Macao, China, August
  10-16, 2019}, pages 5429--5435. ijcai.org.

\bibitem[{Zhu et~al.(2017)Zhu, Xie, Liu, and Sun}]{zhu2017iterative}
Hao Zhu, Ruobing Xie, Zhiyuan Liu, and Maosong Sun. 2017.
\newblock \href {https://doi.org/10.24963/ijcai.2017/595} {Iterative entity
  alignment via joint knowledge embeddings}.
\newblock In \emph{Proceedings of the Twenty-Sixth International Joint
  Conference on Artificial Intelligence, {IJCAI} 2017, Melbourne, Australia,
  August 19-25, 2017}, pages 4258--4264. ijcai.org.

\bibitem[{Zhu et~al.(2019)Zhu, Zhou, Wu, Tan, and Guo}]{zhu2019neighborhood}
Qiannan Zhu, Xiaofei Zhou, Jia Wu, Jianlong Tan, and Li~Guo. 2019.
\newblock \href {https://doi.org/10.24963/ijcai.2019/269} {Neighborhood-aware
  attentional representation for multilingual knowledge graphs}.
\newblock In \emph{Proceedings of the Twenty-Eighth International Joint
  Conference on Artificial Intelligence, {IJCAI} 2019, Macao, China, August
  10-16, 2019}, pages 1943--1949. ijcai.org.

\end{thebibliography}
\bibliographystyle{acl_natbib}


\end{document}